\documentclass{article}

\PassOptionsToPackage{numbers, compress}{natbib}

\usepackage[preprint]{neurips_2026}

\usepackage[utf8]{inputenc} % allow utf-8 input
\usepackage[T1]{fontenc}    % use 8-bit T1 fonts
\usepackage{hyperref}       % hyperlinks
\usepackage{url}            % simple URL typesetting
\usepackage{booktabs}       % professional-quality tables
\usepackage{amsfonts}       % blackboard math symbols
\usepackage{nicefrac}       % compact symbols for 1/2, etc.
\usepackage{microtype}      % microtypography
\usepackage{xcolor}         % colors

\usepackage{wrapfig}
\usepackage{graphicx}
\usepackage{xcolor}
\usepackage[most]{tcolorbox}
\usepackage{graphicx}
\usepackage{multirow}

\usepackage{graphicx}
\usepackage{subcaption}

\newtcolorbox{windowbox}[2][]{%
  enhanced,
  breakable,
  colback=white,
  colframe=black,
  boxrule=0.8pt,
  arc=2mm,
  left=3mm,right=3mm,top=2mm,bottom=2mm,
  fonttitle=\bfseries,
  coltitle=white,
  colbacktitle=black,
  title={#2},
  attach boxed title to top left={xshift=0mm,yshift=-1mm},
  boxed title style={sharp corners, boxrule=0pt, interior style={fill=black}},
  #1
}

\newtcolorbox{fieldbox}[2][]{%
  enhanced,
  breakable,
  colback=gray!3,
  colframe=black!25,
  boxrule=0.5pt,
  arc=1.5mm,
  left=2mm,right=2mm,top=1mm,bottom=1mm,
  fonttitle=\bfseries,
  title={#2},
  listing only,
  listing options={
    basicstyle=\ttfamily\footnotesize,
    columns=fullflexible,
    breaklines=true,
    showstringspaces=false,
  },
  #1
}

\newcommand*{\benchmarkname}{3D-Fit}
\title{Do Language Models Dream of Binding Molecules? \\ Benchmarking LLMs under Spatial Constraints}

\author{%
  \bfseries Thomas MacDougall$^{1}$\quad
  Maksim Kuznetsov$^{1}$ \quad
  Roman Schutski$^{3}$ \quad
  Rim Shayakhmetov$^{2}$\\
  \bfseries Maxim Malkov$^{2}$ \quad
  Vladimir Aladinskiy$^{2}$ \quad
  Alex Aliper$^{2}$ \quad
  Alex Zhavoronkov$^{1,2,3}$ \\[0.6em]
  \small $^{1}$Insilico Medicine Canada Inc. \quad $^{2}$Insilico Medicine AI Limited \quad $^{3}$Insilico Medicine Hong Kong Ltd.\\[0.3em]
}

\begin{document}

\maketitle

\begin{abstract}
Structure-based drug design (SBDD) leverages the 3D structure of protein targets, often complemented by other spatial constraints, to generate candidate binding molecules. While diffusion models have dominated as a leading paradigm for high-quality 3D molecule generation, LLM-based methods are rapidly emerging in molecular design and have shown competitive performance in pocket-conditioned molecular generation. However, their ability to reason about physics and 3D spatial environments is largely underexplored. In this work, we systematically analyze whether current general-purpose LLMs are capable of navigating complex 3D constraints compared to established baselines such as specialized diffusion models. We consider 3D ligand generation conditioned on protein pockets together with ligand- and interaction-derived spatial constraints, including anchor fragments, pharmacophore points, and mandatory pocket-ligand interactions. To enable this evaluation, we introduce \benchmarkname{} -- a token-efficient benchmarking strategy for assessing LLM performance on multi-conditioned spatial molecule generation. Our findings reveal a clear pattern in LLM spatial capabilities: while they still lag behind state-of-the-art approaches, they are promising and can handle multiple spatial constraints simultaneously, enabling scaling to heterogeneous setups. 
\end{abstract}

\section{Introduction}

Designing molecules that satisfy strict spatial constraints is a challenging task in computational chemistry. In realistic drug discovery scenarios, 3D generative models should produce molecular structures simultaneously satisfying multiple heterogeneous spatial requirements.

While specialized diffusion models \cite{peng2022pockettomol, guan2023targetdiff, schneuing2024diffsbdd, huang2024pmdm, gu2024alidiff,  gong2025sgediff, zhang2026sefmol} have demonstrated leading performance in standard pocket-conditioned molecular generation, adapting them to simultaneously handle multiple heterogeneous spatial constraints, although feasible \cite{xie2024diffdec, huang2024ipdiff, ziv2025molsnapper, sako2026diffpharma}, remains non-trivial. Balancing these diverse condition types within a single generative process is challenging, as constraints may differ in scale, rigidity, and importance, and may even conflict with one another. As a result, effective multi-constraint generation therefore requires careful model design, accurate condition encoding, and robust training or inference.

Alternatively, Large Language Models (LLMs) have achieved remarkable success across various computational chemistry and drug discovery tasks \cite{Bhattacharyaetal2024, liu2024chatdrug, yu2024llasmol}, largely owing to their innate capacity to seamlessly process complex task instructions and handle multiple constraints simultaneously. Despite these advances, their ability to explicitly reason about physics and 3D  environments remains largely underexplored. Within the specific domain of 3D molecular design, a handful of recent approaches—such as XYZTransformer \cite{flamshepherd2023xyz}, BindGPT \cite{zholus2024bindgpt}, and nach0-pc \cite{kuznetsov2024nachopc}—have trained specialized language models using textual formulations of 3D structures, relying on a textual description of the spatial environment and/or generated structure. Nevertheless, it remains systematically unverified whether general-purpose LLMs possess the capability to navigate complex 3D molecular design.

The main goal of this work is to assess 3D capabilities of the current state-of-the-art LLMs, whether these models can understand, reason about, and accurately generate ligands that satisfy wide sets of geometrically grounded structural constraints. This benchmark is designed to move beyond pocket-conditioned ligand generation by introducing a richer set of structurally explicit generation conditions (e.g., specific interaction patterns, pharmacophore features, and fragment anchoring), reflecting the criteria medicinal chemists routinely consider when forming molecular hypotheses. Here, we broaden the conditioning space to multiple properties and build comprehensive synthetic benchmark datasets of 3D conditions based on the test complexes from \texttt{CrossDocked2020}~\cite{francoeur2020crossdocked} and \texttt{PLINDER}~\cite{durairaj2024plinder}.

In this work, we propose the \benchmarkname{} benchmark to evaluate 3D molecular generation under a variety of spatial  conditions. The benchmark evaluates protein pocket-only and single-constraint protein pocket-conditioned generation (with anchor fragments or pharmacophore points) to compare state-of-the-art diffusion methods with recent proprietary and open-weight foundation LLMs; also it covers multi-constraint protein pocket-conditioned generation to analyze LLMs success rates and failure modes. To make such evaluation possible, we design spatial condition representations, a structured molecular output format, and a prompting protocol tailored to 3D molecular generation. Finally, we provide a detailed analysis of model performance, including success cases, common failure modes, and the challenges faced by LLMs under increasingly complex spatial constraints.

\section{Related Work}

\paragraph{Structure-based Molecular Benchmarks}
Modern structure-based generative models are based on generated datasets such as CrossDocked2020\cite{francoeur2020crossdocked}, and experimental repositories of structures such as PDBbind\cite{wang2005pdbbind} and Binding MOAD\cite{wagle2023sunsetting}, which provide curated binding protein-ligand complexes. Building on these sources, several benchmarks were proposed: CBGBench\cite{lin2024cbgbench} uses CrossDocked-style data to evaluate de novo generation, linker design, and scaffold hopping through metrics focusing on interaction geometry and substructure validity (mostly 2D); POKMOL-3D\cite{liu2024good} presents $32$ protein targets to benchmark pocket-conditioned 3D generation using active molecule conformations; Durian\cite{nie2024durian} uses experimental affinity and structures to evaluate generative performance; and MolGenBench provides a large-scale evaluation of target-aware lead optimization and de novo design across $120$ targets using $220{,}005$ validated active molecules with some of the tasks having 3D. Our benchmark and dataset differ from the mentioned projects by providing a comprehensive set of multiple spatial conditions for every structure, sheer size of the data and a collection of robust computational metrics.

\paragraph{Pocket-conditioned diffusion models} Pocket-conditioned diffusion models are the leading approaches in structure-based drug discovery at the time of writing. Foundational works such as TargetDiff\cite{guan2023targetdiff} and MolDiff\cite{peng2023moldiff} introduced SE(3)-equivariant diffusion processes to jointly generate molecular coordinates and atom types. Building on these, several works proposed the methods to enhance binding accuracy, like PMDM\cite{huang2024pmdm} which generates the joint distribution of protein and ligand atoms or IPDiff\cite{huang2024ipdiff} which introduces explicit interaction priors to guide the sampling process. DiffSBDD\cite{schneuing2024diffsbdd} introduced scalar property conditioning to pocket-constrained generation, followed by DiffBP\cite{lin2025diffbp} that achieved SOTA accuracy by targeting the 2D properties of druglikeness and QED. Other notable enhancements in pocket-conditioned generation include incorporation of binding-aware features (BindDM\cite{huang2024binddm}). Lastly, SeFMol\cite{zhang2026sefmol} employed reinforcement learning to guide the diffusion process toward high-affinity candidates.

\paragraph{Pocket‑conditioned diffusion with auxiliary conditioning}
Following the success of the diffusion-based models in pocket-conditioned generation, a series of works introduced auxiliary geometric conditions to enhance their practical applicability in drug design. The simplest form of such conditions are rigid \textbf{anchor fragments}. DiffDec\cite{xie2024diffdec} was one of the first models to introduce a preservation mechanism to generate new molecular components around fixed scaffolds. The recent FDC-Diff\cite{chen2026fdcdiff} achieved SOTA results in fragment-to-lead optimization.
\textbf{Pharmacophore points} provide a more general description of spatial patterns matching a range of atoms or atomic groups.
Recently, MolSnapper\cite{ziv2025molsnapper} achieved precise 3D pharmacophore matching during pocket-conditioned generation. 
Another way to control spatial generation is to specify \textbf{mandatory pocket-ligand interaction} priors. IPDiff \cite{huang2024ipdiff} was one of the first diffusion models introducing this condition to pocket-aware generative process. 
The recent DiffPharma\cite{sako2026diffpharma} combines pharmacophore and mandatory pocket-ligand interaction conditioning for even better control of molecular generation.
\begin{figure}[t]
\centering
\includegraphics[width=1\linewidth]{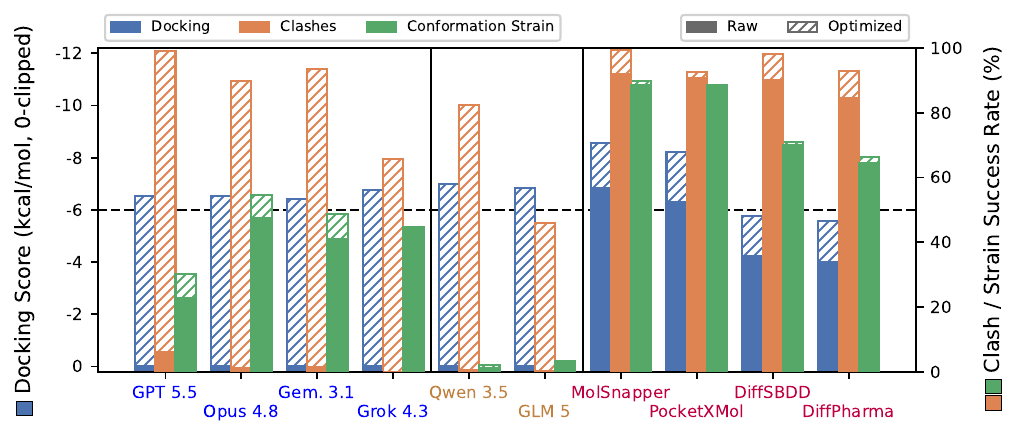}
\caption{\benchmarkname{} benchmark selected results for pocket-only conditioned generation on CrossDocked.}
\label{fig:scores_barplot}
\vspace{-10pt}
\end{figure}

\section{Spatial Condition Descriptions}

\textbf{Protein Pockets} are localized regions of a target protein structure that form cavities capable of accommodating a ligand structure. Pockets are defined by the subset of residues whose atoms line the cavity and determine its geometry and physicochemical environment. In practice, a protein pocket is usually determined by the residues within a given distance cutoff point from a ligand.  

\textbf{Mandatory Protein–Ligand Interactions} are a set of key binding contacts that a candidate ligand is expected to reproduce with certain residues of the target protein pocket. Interaction points are valuable because they provide an interpretable, target-specific description of the binding mode that complements general pocket geometry and can guide molecule generation or evaluation toward ligands that preserve critical contacts with the protein.

\textbf{Anchor Fragments} are chemically significant ligand substructures that are expected to be preserved or approximately reproduced in a generated molecule. These fragments are typically extracted from a reference ligand in its bound conformation and correspond to substructures that contribute to the ligand binding mode. Anchor fragments are useful because they provide a direct way to constrain molecular generation around experimentally or structurally observed ligand geometry while still allowing modifications in the remaining parts of the molecule.

\textbf{Pharmacophore Points} describe the essential spatial arrangement of interaction features responsible for the recognition and binding within a target pocket. They provide an interpretable, target-aware representation of a ligand’s binding mode. While there are protein-based and ligand-based pharmacophore points, we utilize the ligand-based pharmacophores that abstract the molecular structure into key functional features derived primarily from the ligand. By requiring candidate molecules to satisfy these pharmacophore points, the search can be guided toward compounds that maintain critical binding interactions while still allowing the discovery of chemically diverse and novel molecules.
\section{\benchmarkname{}~Benchmark}

Our \benchmarkname{} benchmark is a framework for applying and evaluating multiple 3D condition satisfaction for molecular generation.  It can use any test set of 3D pocket-ligand complexes as condition sources, but in this work we focus on two popular datasets for generative chemistry, \texttt{CrossDocked2020} \cite{francoeur2020crossdocked} and \texttt{PLINDER} \cite{durairaj2024plinder}. Both datasets are well-established sources of test protein-ligand complexes with varying test split strategies and comprehensive specialist model baselines. For each test example, we consider the \textbf{Protein Pocket} as the main 3D condition and up to three additional conditions: \textbf{Mandatory Pocket-Ligand Interaction Points}, \textbf{Anchor Fragments} and \textbf{Pharmacophore Points}. We also propose concise, token-efficient textual descriptions of 3D conditions in the input and a robust textual output format for small molecules in 3D.

\begin{figure}[t]
    \centering
    \includegraphics[width=0.85\textwidth]{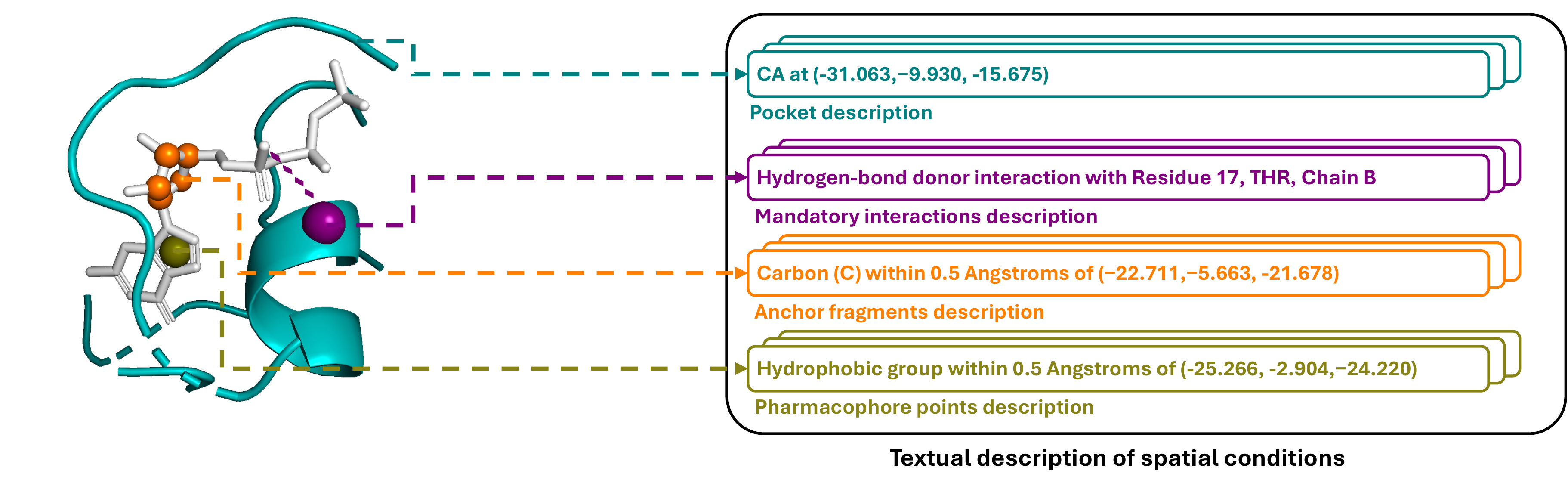}
    \caption{Visualization of spatial conditions and their corresponding textual descriptions.}
    \label{fig:conditions_visualization}
    \vspace{-10pt}
\end{figure}

\subsection{Textual Representation of Spatial Conditions}

In our benchmark, we describe the four spatial conditions using compact textual descriptions that can be provided directly as text to an LLM.  The general  prompt for LLMs benchmarking and examples of the following textual descriptions of spatial conditions are provided in Appendix~\ref{app:templates}.

\paragraph{Protein Pocket} 
Since pockets typically contain hundreds or thousands of atoms, a compact description is crucial for comprehensive yet efficient LLM benchmarking. Although the PDB format \cite{berman2000pdb} is a versatile container for protein structures, it includes substantial redundancy (headers, repeated fields, occupancy/B-factors, element annotations, alternate locations, etc.), which results in unnecessary large token consumption for LLM-based pipelines.

In our work, we propose a textual representation designed to optimize the length of textual protein description without the loss of critical information. To preserve the sequential nature of proteins in this representation, amino acids are described sequentially, beginning with a block that specifies the index and three-letter code of amino acid. For example, the third tyrosine residue is represented as \textcolor{teal}{\texttt{‘Residue~$3$,~TYR’}}. Then, each amino acid's atom is described by its name and coordinates, such as \textcolor{teal}{\texttt{‘CB~at~($3.255$,~$-1.106$,~$8.438$)’}}. We limit atom details to just the atom name and coordinates with three digits after the decimal point, because other properties (atom type, valence, connectivity, charge) can be inferred with standard amino acids. All heavy atoms within an amino acid are listed starting from the backbone and proceeding to the side-chain atoms of the residue (\texttt{N\textrightarrow CA\textrightarrow C\textrightarrow $\ldots$\textrightarrow NH1\textrightarrow NH2}). In case if the target protein contains more than one chain, we specify the chain before the block of amino acids' descriptions corresponding to this chain, e.g. \textcolor{teal}{\texttt{‘Chain A:’}}. 

Since the overall geometry is largely determined by heavy-atom backbones and side chains, we work with hydrogen-depleted proteins to further reduce the number of input tokens. 

\paragraph{Mandatory Pocket-Ligand Interactions} 
Each requirement specifies an interaction type and the corresponding pocket residue (chain ID, residue index, and residue name). For example, an interaction constraint can be expressed as: 

\textcolor{violet}{\texttt{`Hydrophobic interaction with Residue~$13$,~GLU,~Chain~A'}}.

\paragraph{Anchor Fragments} 
We describe anchor fragments in a compact form as a set of \emph{anchor atoms}. Each anchor atom is defined by (i) its atom type and (ii) a spherical tolerance region that indicates where this atom may be placed, represented by the sphere center and the radius. This representation provides a concise way to constrain ligand placement while allowing small positional variability.

For example, to specify a carbon anchor atom  within $0.5$~\AA\ of a given point, we describe it as: 

\textcolor{orange}{\texttt{`Carbon (C) within $0.5$ Angstroms of ($-16.191$,~$-11.325$,~$8.531$)'}}.

\paragraph{Pharmacophore Points} 
As a textual representation, we specify pharmacophore points by type and their 3D positions. If the pharmacophore point has an associated directionality, such as a donor point that should be oriented toward a protein-pocket residue, we do not specify this direction explicitly. Instead, we allow the model to infer the appropriate orientation from the protein pocket description. For example, to specify a hydrogen-bond acceptor that must be positioned within $0.5$~\AA\ of a given coordinate, we provide the LLM with the following instruction:

\textcolor{olive}{\texttt{`Hydrogen bond acceptor within $0.5$ Angstroms of ($22.710$,~$32.862$,~$-24.262$)'}}.

\subsection{Textual Representation of 3D Molecule Output}

The choice of the output format for the generated 3D ligand is essential for our benchmark. The format should be familiar to the LLM's internal knowledge base, and similar to condition representations. It should also be non-redundant and robust, since unnecessary components and format complexity slow down generation and increase the chance of errors that may render the entire output invalid. 

\begin{figure}[t]
\centering
\includegraphics[width=\linewidth]{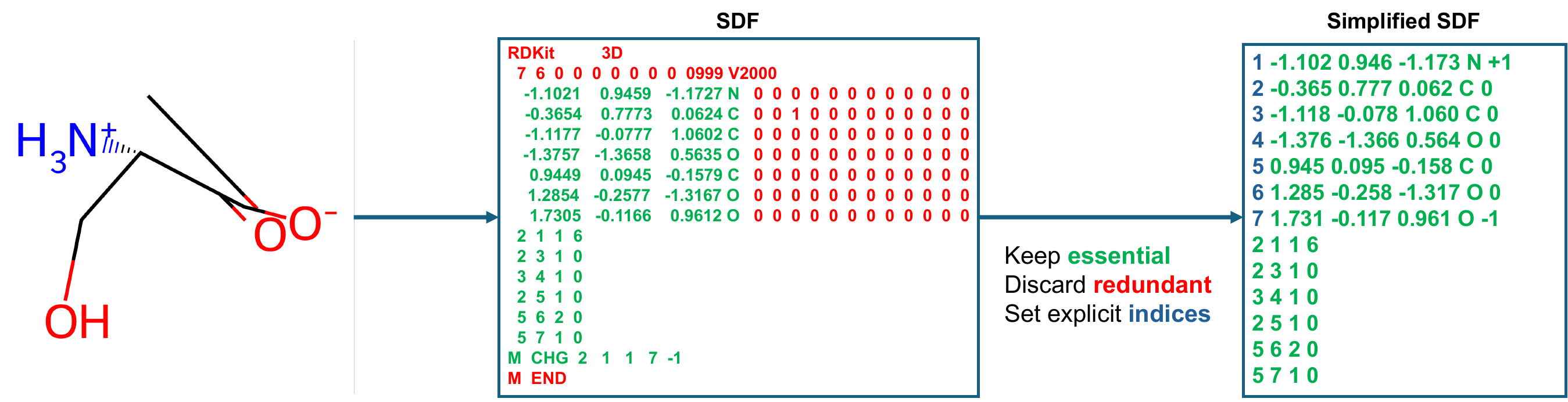}
\caption{Zwitterionic form of Serine amino acid as SDF and Simplified SDF.}
\label{fig:simplified_sdf}
\vspace{-10pt}
\end{figure}

In our work, we propose Simplified SDF (see Fig.~\ref{fig:simplified_sdf}), a modified version of the standard SDF \cite{Dalby1992} designed to reduce the number of textual tokens and format fragility by retaining only key information needed for accurate molecule reconstruction. This format first describes the index, symbol, coordinates and charges of each atom, and then specifies molecular connectivity by listing bonded atom pairs together with their bond types. 

As an ablation study, we also evaluated the SMILES+XYZ format used in \cite{zholus2024bindgpt}\cite{livne2024nacho} as another token-efficient representation (see Appendix~\ref{app:smiles_xyz}). We found that the Simplified SDF consistently outperforms this alternative on nearly all metrics for general-purpose LLMs.  We hypothesize that this is because the benchmarked LLMs were trained on corpora containing data similar to the SDF/Simplified SDF family, as well as the ability to specify atom coordinate before specifying their connectivity.

\subsection{Test Datasets and Conditions Preparation}

We compare the performance of each model on two test datasets, \texttt{CrossDocked2020} v1.3 and \texttt{PLINDER} 2024-06/v2. Each dataset was processed according to five principles: 

\textbf{(i) Literature-consistent test split:} For \texttt{CrossDocked2020}, we used the test split $0$ from the downsampled version of the dataset. For \texttt{PLINDER}, we also used the provided test split.
    
\textbf{(ii) Valid molecule loading:} For both datasets, complexes were included if both the protein and the ligand are valid, i.e. they could be loaded with ProDy \cite{bakan2011prody} and RDKit \cite{landrum2023rdkit} without errors.
    
\textbf{(iii) Binding quality:} For \texttt{CrossDocked2020}, we used complexes with the distance of RMSD <~$0.5$~\AA~ and Vina \cite{trott2010vina} score <~$-6$. For \texttt{PLINDER}, we used all test set complexes as they are.

\textbf{(iv) Drug-like molecule size:} In both datasets, we filtered ligands to retain only those with more than $25$ heavy atoms to better reflect drug-like molecules in the final test sets. We did not apply additional drug-likeness filters to avoid further biasing the data distribution given our focus on 3D conditions.

\textbf{(v) Protein pocket cutoff:} All pockets were identified from full protein structures by selecting all atoms for every residue with at least one atom within $10$~\AA~of the reference ligand center.

The final test sets consist of $948$ complexes for \texttt{CrossDocked2020} and $505$ complexes for \texttt{PLINDER}. For each test example in both datasets, 3D spatial conditions are sampled according to the following processes. The concept of the \benchmarkname{} framework allows for additional sampling or runtime sampling of other protein-ligand complexes for future work in applications such as Reinforcement Learning, as long as the test complexes are not used for any training. For our reported experiments, the sampling was pre-computed and fixed for each example to maintain consistent comparison across models.

\textbf{Protein Pocket} forms the core conditional identity for each test example. For each test object, the pocket-ligand complex is randomly shifted by up to 50~\AA~ in each axis, to prevent any potential coordinate memorization by LLMs in the case of the test set leakage.

\textbf{Mandatory Pocket-Ligand Interactions} are extracted from the reference protein-ligand complex with the ProLIF library~\cite{bouysset2021prolif}. Using the Prolif package, each protein/ligand complex is annotated with all important binding interactions according to their type and residue. One random interaction from the annotated list is sampled and used as a necessary 3D condition. We consider the following interaction types: hydrophobic contacts; hydrogen bonds; $\pi$--$\pi$ stacking; van der Waals contacts; ionic interactions involving anionic or cationic groups; and cation--$\pi$ / $\pi$--cation interactions.

\textbf{Anchor Fragments} are extracted from the ground-truth ligand pose in the holo complex. The scope of potentially chemically important fragments is large, ranging from small functional groups to large scaffold structure. We focus on relatively small anchor fragments, to evaluate fragments as a 3D condition but to allow diversity in the molecule output. For each reference ligand in its binding conformation, we split the molecule according to the BRICS fragmentation algorithm \cite{degen2008brics} and randomly select a fragment with between $3$ and $8$ heavy atoms.  If no fragment meets this criterion, we instead fragment the molecule at all rotatable bonds and select within the same size range. If none are found, we choose a random fragment from the original BRICS decomposition.

\textbf{Pharmacophore Point} is extracted using the Pmapper \cite{kutlushina2018pmapper} package. We annotate each ligand with all possible pharmacophore points independently of the pocket. While it is possible to have multiple points, we sample a single random point from the annotated list to use as a condition. We consider the following pharmacophore types: hydrogen-bond donors, hydrogen-bond acceptors, hydrophobic groups, positive ionizable groups, negative ionizable groups, aromatic groups, and exclusion regions. 

\begin{table*}[t]
\centering
% Conditions: pocket
{\setlength{\tabcolsep}{2pt}
\begin{tabular}{l|cccc|cccc|cc|cc|cc|cccc}
\toprule
\multirow{4}{*}{Model} & \multicolumn{8}{c|}{\textbf{\textcolor{teal}{Pocket}}} & \multicolumn{10}{c}{\textbf{\textcolor{magenta}{Structure}}} \\
\cline{2-19}
 & \multicolumn{4}{c|}{UniDock} & \multicolumn{4}{c|}{PB Inter.} & \multicolumn{2}{c|}{Parsed} & \multicolumn{2}{c|}{RO5} & \multicolumn{2}{c|}{N Hvy} & \multicolumn{4}{c}{PB Intra.} \\
 \cline{2-19}
 & \multicolumn{2}{c}{\texttt{CD}} & \multicolumn{2}{c|}{\texttt{PL}} & \multicolumn{2}{c}{\texttt{CD}} & \multicolumn{2}{c|}{\texttt{PL}} & \multirow{2}{*}{\texttt{CD}} & \multirow{2}{*}{\texttt{PL}} & \multirow{2}{*}{\texttt{CD}} & \multirow{2}{*}{\texttt{PL}} & \multirow{2}{*}{\texttt{CD}} & \multirow{2}{*}{\texttt{PL}} & \multicolumn{2}{c}{\texttt{CD}} & \multicolumn{2}{c}{\texttt{PL}} \\
 & \texttt{R} & \texttt{O} & \texttt{R} & \texttt{O} & \texttt{R} & \texttt{O} & \texttt{R} & \texttt{O} &  &  &  &  &  &  & \texttt{R} & \texttt{O} & \texttt{R} & \texttt{O} \\
\midrule
\textcolor{blue}{GPT 5.5} & \textcolor{red}{19.} & -6.5 & \textcolor{red}{20.} & -6.1 & 6 & \underline{99} & 9 & \textbf{100} & \underline{99} & \textbf{100} & \textbf{99} & \textbf{100} & 23. & 23. & 23 & 30 & 31 & 37\\
\textcolor{blue}{GPT 5.4} & \textcolor{red}{127.} & -6.8 & \textcolor{red}{125.} & -6.4 & 0 & 65 & 0 & 66 & 66 & 67 & 49 & 43 & 29. & 28. & 3 & 4 & 2 & 3 \\
\textcolor{blue}{Opus 4.8} & \textcolor{red}{56.} & -6.5 & \textcolor{red}{53.} & -6.2 & 1 & 90 & 4 & 95 & 90 & 96 & 89 & 95 & 21. & 21. & 47 & 55 & 53 & 58 \\
\textcolor{blue}{Opus 4.7} & \textcolor{red}{52.} & -6.5 & \textcolor{red}{52.} & -6.1 & 2 & 96 & 3 & 95 & 96 & 95 & 95 & 94 & 21. & 21. & 55 & 66 & 57 & 62 \\
\textcolor{blue}{Opus 4.6} & \textcolor{red}{38.} & -6.9 & \textcolor{red}{33.} & -6.5 & 2 & 58 & 5 & 62 & 58 & 62 & 58 & 62 & 22. & 22. & 27 & 31 & 29 & 35 \\
\textcolor{blue}{Sonnet 4.6} & \textcolor{red}{96.} & -6.8 & \textcolor{red}{96.} & -6.4 & 1 & 66 & 0 & 64 & 66 & 64 & 65 & 64 & 23. & 23. & 11 & 12 & 9 & 11 \\
\textcolor{blue}{Gem. 3.1} & \textcolor{red}{57.} & -6.4 & \textcolor{red}{51.} & \textcolor{red}{-6.0} & 1 & 93 & 4 & 94 & 94 & 95 & 85 & 90 & 23. & 23. & 41 & 49 & 41 & 48 \\
\textcolor{blue}{Grok 4.1} & \textcolor{red}{122.} & -6.7 & 
\textcolor{red}{110.} & -6.2 & 0 & 21 & 0 & 25 & 21 & 26 & 21 & 25 & 22. & 22. & 10 & 10 & 10 & 10 \\
\textcolor{blue}{Grok 4.3} & \textcolor{red}{101.} & -6.8 & \textcolor{red}{114.} & -6.2 & 0 & 66 & 0 & 66 & 66 & 66 & 64 & 64 & 21. & 21. & 45 & 45 & 46 & 46 \\
\midrule
\textcolor{brown}{Qwen 3.5} & \textcolor{red}{89.} & -7.0 & \textcolor{red}{88.} & -6.6 & 1 & 82 & 1 & 81 & 83 & 82 & 79 & 77 & 25. & 25. & 1 & 2 & 2 & 3 \\
\textcolor{brown}{GLM 5} & \textcolor{red}{108.} & -6.9 & 
\textcolor{red}{106.} & -6.3 & 0 & 46 & 0 & 43 & 46 & 43 & 44 & 40 & 22. & 23. & 3 & 3 & 3 & 4 \\
\midrule
\textcolor{purple}{PocketXMol} & \underline{-6.3} & \underline{-8.2} & \textcolor{red}{\underline{-4.9}} & -7.1 & \underline{91} & 93 & \textbf{93} & 94 & 93 & 94 & 56 & 55 & 29. & 27. & \textbf{89} & \underline{88} & \textbf{91} & \textbf{90} \\
\textcolor{purple}{DiffSBDD} & \textcolor{red}{-4.2} & \textcolor{red}{-5.8} & \textcolor{red}{-3.5} & \textcolor{red}{-4.9} & 90 & 98 & \underline{91} & \underline{99} & \underline{99} & \underline{99} & 88 & 92 & 17.& 15. & 70 & 71 & 73 & 74 \\
\textcolor{purple}{PDMD} & \textcolor{red}{2.1} & -7.7 & \textcolor{red}{-1.4} & \underline{-7.3} & 1 & 87 & 1 & 86 & 88 & 87 & 58 & 49 & 31. & 33. & \underline{72} & 73 & 70 & 71 \\
\textcolor{purple}{DiffPharma} & \textcolor{red}{-4.0} & \textcolor{red}{-5.6} & \textcolor{red}{-3.2} & \textcolor{red}{-4.6} & 84 & 93 & 89 & 95 & 94 & 95 & 82 & 86 & 17. & 15. & 64 & 66 & 68 & 69 \\
\textcolor{purple}{MolSnapper} & \textbf{-6.8} & \textbf{-8.6} & \textbf{-6.2} & \textbf{-8.2} & \textbf{92} & \underline{99} & \underline{91} & 98 & \textbf{100} & \textbf{100} & 86 & 66 & 27. & 33. & \textbf{89} & \textbf{90} & \underline{85} & \underline{86} \\
\textcolor{purple}{SeFMol} & \textcolor{red}{-0.2} & -6.2 & \textcolor{red}{-3.3} & \textcolor{red}{-5.4} & 0 & \textbf{100} & 1 & \textbf{100} & \textbf{100} & \textbf{100} & \underline{98} & \underline{99} & 17. & 16. & 51 & 52 & 55 & 55 \\
\textcolor{purple}{IPDiff} & - & - & \textcolor{red}{-2.9} & \textcolor{red}{-5.5} & - & - & 1 & \textbf{100} & - & \textbf{100} & - & 78 & - & 17. & - & - & 54 & 57 \\
\textcolor{purple}{BindDM} & \textcolor{red}{-2.9} & \textcolor{red}{-5.3} & \textcolor{red}{-1.9} & \textcolor{red}{-4.4} & 0 & \underline{99} & 1 & \underline{99} & \textbf{100} & \textbf{100} & 73 & 72 & 19. & 17. &43 & 50 & 44 & 53 \\
\bottomrule
\end{tabular}
}
\caption{Benchmarking Results for Pocket-Only Conditioned Generation.}
\label{tab:pocket}
\end{table*}

\newpage
\section{Experiments}

In this section, we benchmark recent closed- and open-weight large language models, together with diffusion models, and analyze their success and failure modes. The modular design of \benchmarkname{} allows conditions to be added or removed from the evaluation depending on the model's capabilities.

\subsection{LLM Models} 

In this work, we use the following recently released LLMs as baselines:

\textcolor{blue}{Proprietary Closed-Weight LLMs}: GPT 5.5~\cite{openai2026gpt55}; GPT 5.4~\cite{openai2026gpt54}; Claude 4.7 Opus~\cite{anthropic2026opus47}; Claude 4.6 Opus~\cite{anthropic2026opus46}; Claude 4.6 Sonnet~\cite{anthropic2026sonnet46}; Gemini 3.1 Pro~\cite{google2026gemini31pro}; Grok 4.1 Fast Reasoning~\cite{xai2025grok41}. 

\textcolor{brown}{Open-Weight LLMs}: Qwen-3.5 (397b-a17b)~\cite{qwen2026qwen35}; DeepSeek v3.2~\cite{deepseekai2025deepseekv32}; GLM-5~\cite{glm5team2026glm5}.

During benchmarking, LLMs were not allowed to perform web searches or use any external tools, ensuring that we evaluated only the knowledge and reasoning capabilities of the models themselves. All LLM results were sampled with a temperature of $1.0$, a top p of $1.0$, maximum new tokens of $8192$ without counting the reasoning tokens, and the default API parameters beyond this. Where it was supported by the model and provider, reasoning effort/reasoning level was set to "high". 

\subsection{Specialist 3D Diffusion Models} 

We also compare against specialist 3D molecular generative models conditioned on various combinations of 3D constraints.  We indicate them with \textcolor{purple}{purple} color. We list the condition sets considered in our benchmark and the corresponding diffusion models supporting them:

\textbf{\textcolor{teal}{Pocket-only}}: PocketXMol~\cite{peng2026pocketxmol}, SeFMol~\cite{zhang2026sefmol}, DiffPharma~\cite{sako2026diffpharma}, MolSnapper~\cite{ziv2025molsnapper}, 
IPDiff~\cite{huang2024ipdiff}, 
BindDM~\cite{huang2024binddm}, DiffSBDD~\cite{schneuing2024diffsbdd} and PMDM~\cite{huang2024pmdm}.

\textbf{\textcolor{teal}{Pocket}+\textcolor{orange}{Anchor fragments}}: PocketXMol~\cite{peng2026pocketxmol}, DiffSBDD~\cite{schneuing2024diffsbdd} and PMDM~\cite{huang2024pmdm}.

\textbf{\textcolor{teal}{Pocket}+\textcolor{olive}{Pharmacophore points}}: DiffPharma~\cite{sako2026diffpharma} and MolSnapper~\cite{ziv2025molsnapper}.

\subsection{Metrics and Evaluation}

\begin{table*}[t]
\centering
\setlength{\tabcolsep}{3pt}
% Conditions: pocket_fragment
{\setlength{\tabcolsep}{2pt}
\begin{tabular}{l|cccc|cccc|cccc|cc|cc|cc|cccc}
\toprule
\multirow{4}{*}{Model} & \multicolumn{8}{c|}{\textbf{\textcolor{teal}{Pocket}}} & \multicolumn{4}{c|}{\textbf{\textcolor{orange}{\shortstack{Anchor}}}} & \multicolumn{10}{c}{\textbf{\textcolor{magenta}{Structure}}} \\
\cline{2-23}
 & \multicolumn{4}{c|}{UniDock} & \multicolumn{4}{c|}{PB Inter.} & \multicolumn{4}{c|}{SR} & \multicolumn{2}{c|}{Parsed} & \multicolumn{2}{c|}{RO5} & \multicolumn{2}{c|}{N Hvy} & \multicolumn{4}{c}{PB Intra.} \\
 \cline{2-23}
 & \multicolumn{2}{c}{\texttt{CD}} & \multicolumn{2}{c|}{\texttt{PL}} & \multicolumn{2}{c}{\texttt{CD}} & \multicolumn{2}{c|}{\texttt{PL}} & \multicolumn{2}{c}{\texttt{CD}} & \multicolumn{2}{c|}{\texttt{PL}} & \multirow{2}{*}{\texttt{CD}} & \multirow{2}{*}{\texttt{PL}} & \multirow{2}{*}{\texttt{CD}} & \multirow{2}{*}{\texttt{PL}} & \multirow{2}{*}{\texttt{CD}} & \multirow{2}{*}{\texttt{PL}} & \multicolumn{2}{c}{\texttt{CD}} & \multicolumn{2}{c}{\texttt{PL}} \\
 & \texttt{R} & \texttt{O} & \texttt{R} & \texttt{O} & \texttt{R} & \texttt{O} & \texttt{R} & \texttt{O} & \texttt{R} & \texttt{O} & \texttt{R} & \texttt{O} &  &  &  &  &  &  & \texttt{R} & \texttt{O} & \texttt{R} & \texttt{O} \\
\midrule
\textcolor{blue}{GPT 5.5} & \textcolor{red}{7.1} & -6.4 & \textcolor{red}{2.9} & \textcolor{red}{-5.9} & 21 & \textbf{99} & 32 & \textbf{99} & \textbf{99} & 0 & \textbf{99} & 1 & \textbf{99} & \textbf{99} & \textbf{99}& \textbf{99} & 22. & 22. & 46 & 55 & 53 & 59 \\
\textcolor{blue}{GPT 5.4} & \textcolor{red}{82.} & -6.6 & \textcolor{red}{61.} & \textcolor{red}{-5.9} & 2 & 66 & 6 & 66 & 67 & 0 & 66 & 0 & 67 & 66 & 54 & 57 & 26. & 26. & 4 & 7 & 4 & 8 \\
\textcolor{blue}{Opus 4.8} & \textcolor{red}{28.} & -6.2 & \textcolor{red}{23.} & \textcolor{red}{-5.7} & 7 & 89 & 13 & 95 & 90 & 1 & \underline{95} & 0 & 90 & 95 & \underline{90} & \underline{95} & 21. & 21. & 50 & 54& 63 & 67 \\
\textcolor{blue}{Opus 4.7} & \textcolor{red}{23.} & -6.1 & \textcolor{red}{15.} & \textcolor{red}{-5.7} & 8 & 91 & 13 & 95 & 91 & 1 & \underline{95} & 0 & 91 & 95 & \underline{90} & 93 & 20. & 20. & 54 & 59 & 65 & 69 \\
\textcolor{blue}{Opus 4.6} & \textcolor{red}{11.} & -6.4 & \textcolor{red}{5.7} & \textcolor{red}{-5.9} & 11 & 74 & 18 & 84 & 74 & 1 & 84 & 2 & 74 & 84 & 74 & 84 & 21. & 21. & 43 & 50 & 56 & 62 \\
\textcolor{blue}{Sonnet 4.6} & \textcolor{red}{33.} & -6.4 & \textcolor{red}{23.} & \textcolor{red}{-5.7} & 5 & 70 & 9 & 74 & 70 & 0 & 74 & 0 & 70 & 74 & 69 & 72 & 21. & 21. & 23 & 26 & 24 & 29 \\
\textcolor{blue}{Gem. 3.1} & \textcolor{red}{39.} & \textcolor{red}{-5.9} & \textcolor{red}{21.} & \textcolor{red}{-5.4} & 7 & \underline{94} & 11 & \underline{96} & 94 & 1 & \underline{95} & 0 & 95 & \underline{96} & 82 & 88 & 22. & 22. & 57 & 64 & 57 & 66 \\
\textcolor{blue}{Grok 4.1} & \textcolor{red}{45.} & -6.9 & \textcolor{red}{29.} & -6.1 & 4 & 17 & 4 & 15 & 17 & 0 & 16 & 0 & 17 & 16 & 16 & 15 & 20. & 20. & 3 & 3 & 2 & 3 \\
\textcolor{blue}{Grok 4.3} & \textcolor{red}{55.} & -6.9 & \textcolor{red}{34.} & -6.3 & 2 & 37 & 4 & 33 & 37 & 0 & 32 & 0 & 38 & 34 & 33 & 29 & 22. & 23. & 6 & 6 & 3 & 3 \\
\midrule
\textcolor{brown}{Qwen 3.5} & \textcolor{red}{59.} & -6.9 & \textcolor{red}{35.} & -6.3 & 1 & 49 & 4 & 47 & 50 & 0 & 47 & 0 & 50 & 48 & 42 & 41 & 27. & 27. & 0 & 0 & 0 & 0 \\
\textcolor{brown}{GLM 5} & \textcolor{red}{43.} & -6.2 & \textcolor{red}{32.} & \textcolor{red}{-5.5} & 2 & 42 & 2 & 47 & 43 & 0 & 47 & 0 & 43 & 47 & 41 & 46 & 21. & 21. & 3 & 4 & 4 & 5 \\
\midrule
\textcolor{purple}{PocketXMol} & \textbf{-6.3} & \underline{-7.9} & \textcolor{red}{\textbf{-4.9}} & \underline{-7.0} & \textbf{87} & 90 & \textbf{76} & 85 & 90 & \underline{15} & 86 & \underline{9} & 90 & 86 & 59 & 30& 27. & 34. & \textbf{83} & \textbf{80} & \underline{72} & \underline{70} \\
\textcolor{purple}{DiffSBDD} & \textcolor{red}{-2.3} & \textcolor{red}{-4.9} & \textcolor{red}{-1.6} & \textcolor{red}{-4.2} & \underline{68} & 88 & \underline{70} & 93 & \underline{98} & \textbf{18} & \textbf{99} & \textbf{10} & \underline{98} & \textbf{99} & 79 & 78 & 18. & 16. & \underline{78} & \underline{74} & \textbf{73} & \textbf{77} \\
\textcolor{purple}{PDMD} & \textcolor{red}{\underline{-5.4}} & \textbf{-8.5} & \textcolor{red}{\underline{-4.0}} & \textbf{-7.3} & 36 & 58 & 32 & 52 & 59 & 9 & 51 & 4 & 60 & 53 & 36 & 32 & 29. & 27. & 48 & 48 & 44 & 44\\
\bottomrule
\end{tabular}
}
\caption{Benchmarking Results for Pocket+Anchor Fragment Conditioned Generation.}
\label{tab:pocket+fragment}
\end{table*}

We evaluated all generated molecular structures against sampled conditions for each protein-ligand complex. Evaluation of the generated molecules is done on the basis of two key pillars of success; \textbf{3D~Molecular Validity} and \textbf{3D Condition Success}. For binary pass/fail metrics, we report the Success Rate (SR, \%) on all outputs, to not bias results by validity.

\textbf{3D Molecular Validity} metrics evaluate the quality of the generated 3D molecules independently of any external conditions. We first report whether the input molecule can be successfully reconstructed and sanitized (Parsed). We also show the average number of heavy atoms (N Hvy) to assess the size of generated molecules. Along with these metrics, we compute the percentage of molecules that passes Lipinski's rule of five (RO5) to roughly estimate the druglikeness of molecules. We then use PoseBusters~\cite{buttenschoen2024posebusters} to assess 3D plausibility. Specifically, we group the conformation-related PoseBusters checks into a single metric, \textit{PoseBusters Intra-Molecular} (PB Intra), and count a molecule as successful only if it passes all of the following filters: bond lengths, bond angles, internal steric clashes, aromatic ring flatness, non-aromatic ring non-flatness, and double bond flatness.

\textbf{3D Condition Success} metrics assess the molecule's satisfaction of the generation conditions tied to each test example.
We report this metrics for ligand position before (\texttt{R}=Raw) and after (\texttt{O}=Optimized) local ligand pose optimization with UniDock. For Pharmacophores, Anchor Fragments, and Pocket-Ligand Interaction point conditions, we evaluate success from the 3D positioning of atoms/features. If the generated molecule contains the conditioned feature, atom, or interaction in the desired location or with the desired residue, we consider it as a success. We do not evaluate the 2D-connectivity of Anchor Fragments or other features, only 3D placement. For Protein Pockets, evaluating success is more nuanced as there is no obvious pass condition universal for all proteins, so we report two metrics. We calculate and report Unidock \cite{yu2023unidock} Score (Median, kcal/mol, Lower is better). We highlight with \textcolor{red}{red} if the median model Unidock Score higher than $-6$ threshold. We also group the pocket-related PoseBusters checks into a single metric, \textit{PoseBusters Inter-Molecular} (PB Inter), and count a molecule as successful only if it passes all of the following filters: protein-ligand maximum distance, minimum distance to the protein and volume overlap with the protein.

\begin{table*}[t]
\centering
\setlength{\tabcolsep}{3pt}
\resizebox{1\linewidth}{!}{
% Conditions: pocket_pharmacophore
{\setlength{\tabcolsep}{2pt}
\begin{tabular}{l|cccc|cccc|cccc|cc|cc|cc|cccc}
\toprule
\multirow{4}{*}{Model} & \multicolumn{8}{c|}{\textbf{\textcolor{teal}{Pocket}}} & \multicolumn{4}{c|}{\textbf{\textcolor{olive}{Phar.}}} & \multicolumn{10}{c}{\textbf{\textcolor{magenta}{Structure}}} \\
\cline{2-23}
 & \multicolumn{4}{c|}{UniDock} & \multicolumn{4}{c|}{PB Inter.} & \multicolumn{4}{c|}{SR} & \multicolumn{2}{c|}{Parsed} & \multicolumn{2}{c|}{RO5} & \multicolumn{2}{c|}{N Hvy} & \multicolumn{4}{c}{PB Intra.} \\
 \cline{2-23}
 & \multicolumn{2}{c}{\texttt{CD}} & \multicolumn{2}{c|}{\texttt{PL}} & \multicolumn{2}{c}{\texttt{CD}} & \multicolumn{2}{c|}{\texttt{PL}} & \multicolumn{2}{c}{\texttt{CD}} & \multicolumn{2}{c|}{\texttt{PL}} & \multirow{2}{*}{\texttt{CD}} & \multirow{2}{*}{\texttt{PL}} & \multirow{2}{*}{\texttt{CD}} & \multirow{2}{*}{\texttt{PL}} & \multirow{2}{*}{\texttt{CD}} & \multirow{2}{*}{\texttt{PL}} & \multicolumn{2}{c}{\texttt{CD}} & \multicolumn{2}{c}{\texttt{PL}} \\
 & \texttt{R} & \texttt{O} & \texttt{R} & \texttt{O} & \texttt{R} & \texttt{O} & \texttt{R} & \texttt{O} & \texttt{R} & \texttt{O} & \texttt{R} & \texttt{O} &  &  &  &  &  &  & \texttt{R} & \texttt{O} & \texttt{R} & \texttt{O} \\
\midrule
\textcolor{blue}{GPT 5.5} & \textcolor{red}{12.} & -6.4 & \textcolor{red}{11.} & \textcolor{red}{-5.9} & 10 & \textbf{99} & 16 & \textbf{99} & \textbf{99} & \textbf{99} & \textbf{100} & \textbf{99} & \underline{99} & \textbf{100} & \textbf{99} & \textbf{99} & 22. & 22. & 32 & 41 & 29 & 39 \\
\textcolor{blue}{GPT 5.4} & \textcolor{red}{109.} & -6.6 & \textcolor{red}{108.} & -6.2 & 1 & 67 & 2 & 65 & 68 & 68 & 66 & 66 & 69 & 66 & 48 & 46 & 28. & 28. & 5 & 9 & 7 & 9 \\
\textcolor{blue}{Opus 4.8} & \textcolor{red}{52.} & -6.5 & \textcolor{red}{41.} & -6.1 & 2 & 84 & 5 & 90 & 85 & 85 & 90 & 90 & 85 & 90 & 85 & 90 & 21. & 21. & 42 & 48 & 44 & 50 \\
\textcolor{blue}{Opus 4.7} & \textcolor{red}{43.} & -6.4 & \textcolor{red}{43.} & \textcolor{red}{-6.0} & 4 & 92 & 6 & 94 & 93 & 93 & 94 & \underline{94} & 93 & 94 & \underline{91} & \underline{93} & 20. & 20. & 51 & \underline{62} & 50 & 61 \\
\textcolor{blue}{Opus 4.6} & \textcolor{red}{22.} & -6.8 & \textcolor{red}{23.} & -6.2 & 4 & 60 & 6 & 65 & 60 & 60 & 65 & 65 & 60 & 65 & 59 & 64 & 22. & 22. & 26 & 31 & 29 & 35 \\
\textcolor{blue}{Sonnet 4.6} & \textcolor{red}{65.} & -6.7 & \textcolor{red}{55.} & -6.3 & 2 & 61 & 3 & 65 & 61 & 61 & 65 & 65 & 61 & 65 & 60 & 64 & 22. & 22. & 8 & 11 & 8 & 11 \\
\textcolor{blue}{Gem. 3.1} & \textcolor{red}{48.} & -6.1 & \textcolor{red}{37.} & \textcolor{red}{-5.7} & 4 & \underline{94} & 9 & 94 & \underline{94} & \underline{94} & \underline{95} & \underline{94} & 95 & \underline{95} & 85 & 86 & 23. & 23. & 41 & 51 & 43 & 52 \\
\textcolor{blue}{Grok 4.1} & \textcolor{red}{102.} & \underline{-7.4} & \textcolor{red}{90.} & -6.6 & 0 & 15 & 0 & 15 & 15 & 15 & 16 & 16 & 16 & 16 & 13 & 14 & 24. & 24. & 1 & 1 & 2 & 2 \\
\textcolor{blue}{Grok 4.3} & \textcolor{red}{74.} & -6.7 & \textcolor{red}{70.} & -6.3 & 1 & 52 & 3 & 54 & 51 & 51 & 53 & 53 & 53 & 55 & 50 & 51 & 22. & 22. & 31 & 31 & 30 & 31 \\
\midrule
\textcolor{brown}{Qwen 3.5} & \textcolor{red}{89.} & \underline{-7.4} & \textcolor{red}{78.} & \underline{-6.8} & 1 & 42 & 2 & 45 & 42 & 42 & 45 & 45 & 43 & 45 & 35 & 37 & 26. & 26. & 0 & 0 & 0 & 0 \\
\textcolor{brown}{GLM 5} & \textcolor{red}{64.} & -6.9 & \textcolor{red}{55.} & -6.3 & 1 & 27 & 2 & 33 & 27 & 27 & 33 & 33 & 27 & 33 & 24 & 30 & 22. & 22. & 1 & 1 & 1 & 1 \\
\midrule
\textcolor{purple}{DiffPharma} & \textcolor{red}{\underline{-4.2}} & \textcolor{red}{-5.7} & \textcolor{red}{\underline{-3.3}} & \textcolor{red}{-4.9} & \textbf{88} & \underline{94} & \textbf{89} & \underline{95} & 73 & 73 & 78 & 77 & 95 & \underline{95} & 84 & 87 & 17. & 16. & \underline{53} & 55 & \underline{61} & \underline{62} \\
\textcolor{purple}{MolSnapper} & \textbf{-6.1} & \textbf{-8.1} & \textcolor{red}{\textbf{-4.4}} & \textbf{-7.3} & \underline{84} & \textbf{99} & \underline{82} & \textbf{99} & 91 & 91 & 94 & \underline{94} & \textbf{100} & \textbf{100} & 86 & 64 & 27. & 32. & \textbf{90} & \textbf{91} & \textbf{86} & \textbf{89} \\
\bottomrule
\end{tabular}
}
}
\caption{Benchmarking Results for Pocket+Pharmacophore Point Conditioned Generation.}
\label{tab:pocket+pharmacophore}
\end{table*}

\subsection{Results}

We organize the results according to the following conditioning settings:

\textbf{(i) \textcolor{teal}{Pocket-only} Conditioning} is the most widely supported setting, and diffusion models that support additional constraints typically also support pocket-only conditioning. Table \ref{tab:pocket} shows these results.

\textbf{(ii) \textcolor{teal}{Pocket}+\textcolor{orange}{Fragment} Conditioning} and \textbf{\textcolor{teal}{Pocket}+\textcolor{olive}{Pharmacophore} Conditioning} are supported by only a small number of diffusion models. Tables \ref{tab:pocket+fragment} and \ref{tab:pocket+pharmacophore} show these results, respectively.

\textbf{(iii) \textcolor{teal}{Pocket}+\textcolor{violet}{Mandatory Interaction}+\textcolor{olive}{Pharmacophore}+\textcolor{orange}{Fragment}  Conditioning} is not natively supported by any diffusion or other model we are aware of. Still LLMs are easy to prompt in this setting, so we compare only the LLMs models. Table \ref{tab:all_conditions} shows these results.

\begin{table*}[t]
\centering
\setlength{\tabcolsep}{3pt}
\resizebox{1\linewidth}{!}{
% Conditions: all_conditions
{\setlength{\tabcolsep}{2pt}
\begin{tabular}{l|cccc|cccc|cccc|cccc|cccc|cc|cc|cc|cccc}
\toprule
\multirow{4}{*}{Model} & \multicolumn{8}{c|}{\textbf{\textcolor{teal}{Pocket}}} & \multicolumn{4}{c|}{\textbf{\textcolor{orange}{\shortstack{Anchor}}}} & \multicolumn{4}{c|}{\textbf{\textcolor{olive}{Phar.}}} & \multicolumn{4}{c|}{\textbf{\textcolor{violet}{M.Interact.}}} & \multicolumn{10}{c}{\textbf{\textcolor{magenta}{Structure}}} \\
\cline{2-31}
 & \multicolumn{4}{c|}{UniDock} & \multicolumn{4}{c|}{PB Inter.} & \multicolumn{4}{c|}{SR} & \multicolumn{4}{c|}{SR} & \multicolumn{4}{c|}{SR} & \multicolumn{2}{c|}{Parsed} & \multicolumn{2}{c|}{RO5} & \multicolumn{2}{c|}{N Hvy} & \multicolumn{4}{c}{PB Intra.} \\
 \cline{2-31}
 & \multicolumn{2}{c}{\texttt{CD}} & \multicolumn{2}{c|}{\texttt{PL}} & \multicolumn{2}{c}{\texttt{CD}} & \multicolumn{2}{c|}{\texttt{PL}} & \multicolumn{2}{c}{\texttt{CD}} & \multicolumn{2}{c|}{\texttt{PL}} & \multicolumn{2}{c}{\texttt{CD}} & \multicolumn{2}{c|}{\texttt{PL}} & \multicolumn{2}{c}{\texttt{CD}} & \multicolumn{2}{c|}{\texttt{PL}} & \multirow{2}{*}{\texttt{CD}} & \multirow{2}{*}{\texttt{PL}} & \multirow{2}{*}{\texttt{CD}} & \multirow{2}{*}{\texttt{PL}} & \multirow{2}{*}{\texttt{CD}} & \multirow{2}{*}{\texttt{PL}} & \multicolumn{2}{c}{\texttt{CD}} & \multicolumn{2}{c}{\texttt{PL}} \\
 & \texttt{R} & \texttt{O} & \texttt{R} & \texttt{O} & \texttt{R} & \texttt{O} & \texttt{R} & \texttt{O} & \texttt{R} & \texttt{O} & \texttt{R} & \texttt{O} & \texttt{R} & \texttt{O} & \texttt{R} & \texttt{O} & \texttt{R} & \texttt{O} & \texttt{R} & \texttt{O} &  &  &  &  &  &  & \texttt{R} & \texttt{O} & \texttt{R} & \texttt{O} \\
\midrule
\textcolor{blue}{GPT 5.5} & \textcolor{red}{\textbf{3.0}} & -6.3 & \textcolor{red}{\textbf{1.7}} & \textcolor{red}{-5.7} & \textbf{26} & \textbf{99} & \textbf{37} & \textbf{99} & \textbf{99} & \underline{1} & \textbf{99} & 0 & \textbf{99} & \textbf{99} & \textbf{99} & \textbf{99} & \textbf{73} & \textbf{28} & \textbf{82} & \textbf{28} & \textbf{99} & \textbf{99} & \textbf{98} & \textbf{98} & 21. & 21. & \textbf{41} & \textbf{48} & \textbf{42} & \textbf{48} \\
\textcolor{blue}{GPT 5.4} & \textcolor{red}{68.} & -6.7 & \textcolor{red}{56.} & -6.1 & 1 & 56 & 2 & 59 & 61 & \textbf{2} & 63 & \textbf{2} & 60 & 60 & 61 & 61 & 27 & 17 & 31 & 19 & 61 & 63 & 47 & 50 & 26. & 26. & 2 & 2 & 1 & 2 \\
\textcolor{blue}{Opus 4.8} & \textcolor{red}{11.} & -6.1 & \textcolor{red}{6.4} & \textcolor{red}{-5.5} & 9 & 70 & 15 & 76 & 70 & 0 & 76 & 0 & 70 & 70 & 76 & 76 & 44 & 20 & 52 & 22 & 70 & 76 & 69 & 74 & 20. & 21. & 24 & 26 & 32 & 35 \\
\textcolor{blue}{Opus 4.7} & \textcolor{red}{10.} & \textcolor{red}{-6.0} & \textcolor{red}{5.5} & \textcolor{red}{-5.4} & \underline{11} & 76 & \underline{20} & 84 & 76 & \underline{1} & 84 & 0 & 76 & 76 & 84 & 83 & 37 & 19 & 46 & 21 & 76 & 84 & 73 & 82 & 20. & 20. & 28 & 32 & 36 & 39 \\
\textcolor{blue}{Opus 4.6} & \textcolor{red}{\underline{7.2}} & -6.2 & \textcolor{red}{\underline{5.1}} & \textcolor{red}{-5.6} & 10 & 68 & 17 & 79 & 68 & \underline{1} & 79 & 0 & 68 & 68 & 79 & 79 & 35 & 18 & 41 & 22 & 68 & 79 & 67 & 78 & 21. & 21. & 31 & 34 & \underline{38} & \underline{41} \\
\textcolor{blue}{Sonnet 4.6} & \textcolor{red}{15.} & \textcolor{red}{-5.8} & \textcolor{red}{9.6} & \textcolor{red}{-5.1} & 6 & 60 & 10 & 67 & 61 & 0 & 67 & 0 & 60 & 60 & 66 & 66 & 33 & 16 & 37 & 18 & 61 & 67 & 58 & 64 & 20. & 20. & 17 & 18 & 18 & 21 \\
\textcolor{blue}{Gem. 3.1} & \textcolor{red}{29.} & -6.1 & \textcolor{red}{17.} & \textcolor{red}{-5.5} & 5 & \underline{92} & 9 & \underline{93} & \underline{93} & \underline{1} & \underline{94} & \underline{1} & \underline{93} & \underline{93} & \underline{94} & \underline{94} & \underline{68} & \underline{27} & \underline{73} & \underline{26} & \underline{93} & \underline{95} & \underline{80} & \underline{84} & 23. & 23. & \underline{35} & \underline{39} & 32 & 38 \\
\textcolor{blue}{Grok 4.1} & \textcolor{red}{48.} & \underline{-7.1} & \textcolor{red}{49.} & \textbf{-6.6} & 3 & 18 & 2 & 13 & 19 & 0 & 13 & 0 & 17 & 17 & 13 & 12 & 7 & 4 & 6 & 4 & 19 & 13 & 15 & 12 & 20. & 20. & 3 & 3 & 2 & 2 \\
\textcolor{blue}{Grok 4.3} & \textcolor{red}{43.} & -7.0 & \textcolor{red}{44.} & \underline{-6.5} & 2 & 43 & 2 & 35 & 43 & 0 & 35 & 0 & 43 & 43 & 34 & 34 & 21 & 11 & 18 & 10 & 44 & 36 & 34 & 32 & 23. & 23. & 1 & 1 & 1& 2 \\
\midrule
\textcolor{brown}{Qwen 3.5} & \textcolor{red}{44.} & \textbf{-7.3} & \textcolor{red}{41.} & -6.3 & 1 & 47 & 2 & 48 & 49 & \underline{1} & 49 & 0 & 48 & 47 & 48 & 47 & 22 & 11 & 26 & 11 & 50 & 49 & 36 & 36 & 27. & 27. &0 & 0 & 0 & 0 \\
\textcolor{brown}{GLM 5} & \textcolor{red}{23.} & -6.1 & \textcolor{red}{18.} & \textcolor{red}{-5.5} & 3 & 39 & 5 & 39 & 39 & 0 & 39 & 0 & 39 & 39 & 38 & 38 & 28 & 9 & 31 & 10 & 39 & 39 & 33 & 36 & 21. & 21. & 1 & 1 &1 & 2 \\
\bottomrule
\end{tabular}
}
}
\caption{Results for Pocket+Interaction+Anchor+Pharmacophore Conditioned Generation.}
\label{tab:all_conditions}
\end{table*}

\section{Discussion}
\label{sec:discussion}

In this section, we discuss the key observations from the benchmarking results.

\paragraph{Increasing Spatial Conditions for LLMs}
Across all experiments, our results demonstrate that LLMs show emerging ability to follow spatial constraints, especially when the number of conditions increases. The successful parsing metrics indicate that the models generally understand the required output format. However, the PoseBusters intra-molecular filters (PB Intra) reveal variability in the quality of the generated conformations. LLMs perform particularly well on Anchor Fragment and Pharmacophore Point conditioning, possibly because these conditions are described in a form that closely resembles the corresponding output entries needed to satisfy them. In contrast, the models struggle more with Mandatory Interaction Point condition, which may be due to their more abstract nature and poor understanding of specific protein-ligand interactions.

When additional conditions are provided, the models achieve substantially better Pocket metrics compared to the Pocket-only setting. This suggests that more seed-ligand information helps the models generate better molecules, which begin to fit inside the protein pocket. The results also indicate that the models may perform only limited exploration beyond molecules that directly satisfy the given conditions; therefore, adding more constraints may encourage more accurate exploration of chemical space in order to satisfy all specified requirements.

\paragraph{Docking Scores of Generated Molecules}

Although LLMs can often produce molecules that satisfy some structural constraints after optimization, their raw poses are poor binders (see Fig.~\ref{fig:docked_mols}). Before local UniDock optimization (\texttt{R}), all LLM UniDock scores are above the $-6$ threshold, and are often strongly positive, indicating severe steric clashes with the pocket or highly unfavorable placement. Local UniDock optimization (\texttt{O}) substantially improves these poses, bringing most LLM scores to the moderate range of roughly $-6.0$ to $-7.0$ (see Fig.~\ref{fig:unidock_score_dists}). However, they still remain behind the best pocket-specific diffusion models, which achieve substantially lower optimized scores, e.g. MolSnapper, PocketXMol and PDMD. Diffusion models also produce poses with roughly the same binding mode before and after optimization. See Appendix \ref{app:unidock_scores} for the details. These results suggest that while LLMs generally follow the desired spatial conditions, they often do so by producing poses with non-optimal geometry and fail to produce poses that can be optimized without substantial ligand repositioning. In particular, the generated molecules frequently exhibit external steric clashes with the pocket, as reflected by poor raw UniDock scores and low inter-molecular PoseBusters pass rates, as well as internal structural problems indicated by weaker intra-molecular validity. Thus, satisfying the requested spatial constraints alone is not sufficient for generating strong binders: the molecules must also adopt physically plausible geometries, both with respect to the protein environment and their own internal structure.

\paragraph{Diffusion Models and Intermolecular Filters}

The PoseBusters inter-molecular filters show a different failure mode. For LLMs, raw inter-molecular pass rates are generally very low, consistent with their poor raw UniDock scores, but local optimization can substantially improve both docking scores and inter-molecular validity. Diffusion models are more mixed: several methods obtain high inter-molecular pass rates after optimization, but this does not always correspond to strong UniDock scores. For example, DiffSBDD, DiffPharma, SeFMol, IPDiff, and BindDM can pass many inter-molecular filters while still having relatively weak optimized UniDock scores, whereas MolSnapper combines high inter-molecular validity with the best docking scores. Thus, inter-molecular filter success is useful for detecting physically implausible poses, but it is not by itself a sufficient proxy for poses quality. UniDock scores remain the more direct metric for ranking molecular poses, while PoseBusters filters should be interpreted as complementary validity checks rather than definitive measures of binder quality.

\section{Limitations and Impact}

\paragraph{Limitation}
The lack of statistical significance in our experimental results is a limitation. This decision was made to instead conduct a better analysis of more available models. Another limitation is choosing random fragments as anchor fragments. Other works involving fixed fragments consider important substructures such as scaffolds. We consider fragments as a 3D constraint independent of chemical validity and the protein enviornment. Another limitation is the usage of one pharmacophore point and one mandatory interaction point when our developed sampling strategy supports more.

\paragraph{Broader Impacts}
Potential positive impacts of using off-the-shelf LLMs for drug discovery include reducing the costs required to develop new medicines. At the same time, these capabilities raise dual-use concerns, as widely available models could potentially be misused to generate harmful or toxic compounds. Although our work focuses on evaluation rather than deployment, it highlights the need for misuse-aware filtering, access controls in high-risk settings, and chemical-structure safeguards in future LLM-based molecular design systems.

\section{Conclusion}
\label{sec:conclusion}

In this work, we introduce \benchmarkname{}, a benchmark for evaluating 3D molecular generation under increasingly complex spatial conditions. We also define textual representations for these conditions and propose a structured output format for generated 3D molecules.

Our results show that current frontier LLMs exhibit a meaningful ability to parse spatial constraints instructions and generate molecules that satisfy explicit local 3D constraints. In particular, LLMs perform well when conditions are expressed in a form that can be directly copied, mirrored, or locally reconstructed in the output, such as anchor fragments and pharmacophore points. Moreover, adding more spatial conditions often improves pocket-related metrics, suggesting that additional seed-ligand information helps LLMs place generated molecules more accurately in 3D space. These findings indicate that LLMs possess emerging capabilities for instruction-following in 3D environments.

However, our benchmark also reveals important limitations. Although LLMs often follow the desired spatial conditions, they frequently produce poses with non-optimal geometry. Before local UniDock optimization, their docking scores are consistently poor, often indicating severe steric clashes with the protein pocket. After optimization, the scores improve substantially, but still remain weaker than those of the best diffusion-based generators. LLM-generated molecules also show internal structural issues, as reflected by weaker intra-molecular validity compared to the strongest specialized models. Thus, local constraint satisfaction alone is not sufficient for effective structure-based design: generated molecules must also be physically plausible both internally and in their placement relative to the pocket.

Overall, \benchmarkname{} highlights both the promise and the current shortcomings of off-the-shelf LLMs for 3D molecular design. LLMs are increasingly capable of handling multi-condition generation prompts, but they do not yet match diffusion models in physical plausibility and binding quality. We hope this benchmark will support future work on domain-specific training, improved 3D molecular representations, and more reliable evaluation for multi-constraint structure-based drug design.
\section{Code and Data Availability}

\label{app:code_and_data_availability}

The benchmark's code data are accessible via the following link: \url{https://github.com/insilicomedicine/bench-3d-fit}.

\bibliography{bibliography}
\bibliographystyle{unsrtnat}

%%%%%%%%%%%%%%%%%%%%%%%%%%%%%%%%%%%%%%%%%%%%%%%%%%%%%%%%%%%%

\appendix

\newpage
\section{Benchmarking templates}
\label{app:templates}

\begin{windowbox}[width=\linewidth]{General template for LLM benchmarking}
    \begin{fieldbox}{}
    \small
    Generate the three-dimensional structure of a ligand molecule given the following 3D generation conditions:\\
    \textcolor{teal}{
    The ligand molecule must bind to the following protein pocket, meaning good interactions with the protein pocket atoms and no clashes (atoms with very close coordinates) with the protein pocket atoms:\\
    Atoms are grouped first by chain, then by residue, and are referenced using atom names from PDB-style protein notation.\\
    \{ Pocket description \}}\\
    \textcolor{orange}{
    The ligand molecule must contain the following atoms and their 3D coordinates within 0.5 Angstroms of the given coordinates:\\
    \{ Anchor fragments description \}}\\
    \textcolor{olive}{
    The ligand molecule must have the following pharmacophore points:\\
    \{ Pharmacophore points description \}}\\
    \textcolor{violet}{
    The ligand molecule must have the following interactions with specific residues in the protein pocket:\\
    \{ Mandatory interactions description \}\\
    }
    The most important condition is generating a valid molecule with plausible 2D and 3D structure. This means reasonable bond lengths, angles, and no overlapping with protein pocket.
    If all conditions are not satisfiable simultaneously, prioritize the validity of the generated ligand molecule.\\
    \textcolor{magenta}{
    The output molecule must have a valid 2D and 3D structure enclosed in <sdf> and </sdf> tags.
    In this format, the molecule is provided as a simplified SDF/MolBlock string. Which has two parts, an atom block and a bond block.\\
    The atom block contains a line for each heavy atom (non-hydrogen atom), 6 values per line. First is the atom index, starting from 1, followed 3 values for the x y z coordinates in Angstroms with 3 decimal places each, fifth is the atomic symbol and sixth is the charge on the atom (0 for no charge, -1 for negative charge, +1 for positive charge)\\
    The bond block contains a line for each bond, 4 values per line. First is the start atom index from the atom block, the second is the end atom index from the atom block, the third is the bond order (1 for single, 2 for double, 3 for triple) and fourth is the stereoscopy of the bond (if applicable, 0 for none, 1 for up, 6 for down).\\
    Bonds do not have a unique identifier index, the start and end atom indices are used to identify the bond. Hydrogen atoms are not included in this format and are instead determined implicitly from the available valencies of the listed heavy atoms \\
    \{ Simplified SDF representation example \}}\\
    Important: Before finalizing, explicitly verify the output molecule: format validity, 3d plausibility, all heavy atoms have associated coordinates, and all conditions are satisfied. \\
    If uncertain or unable to satisfy all conditions, prefer a ligand molecule that satistifes the original validity conditions: \\
    The ligand molecule must be a valid molecule, meaning it is a single connected structure with correct valencies for all atoms. \\
    The ligand molecule must have a 3D structure, and the bond lengths and angles must obey chemical rules such that the strain energy of the molecule is minimal. \\
    The ligand molecule should be druglike and fill the pocket of the protein, forming good interactions with the protein pocket atoms without overlapping. \\
    To be druglike, the ligand molecule should satisfy Lipinski's rule of five. \\
    The ligand molecule should have a number of heavy atoms between 20 and 40. \\
    The ligand molecule should have a number of hydrogen bond donors less than 5. \\
    The ligand molecule should have a number of hydrogen bond acceptors less than 10. \\
    The ligand molecule should have a logP value less than 5. \\
    Do not output any molecule unless all checks pass. If you see the problems with unrealistic atom coordinates, stop generating the current molecule, close the tags and generate a new molecule from scratch in a new set of mol tags. Repeat this process as necessary to output a valid molecule. \\
    \end{fieldbox}{}
\end{windowbox}

\begin{windowbox}[width=\linewidth]{Simplified SDF representation example}
\begin{fieldbox}[colback=magenta!3]{}
    \textcolor{magenta}{
    For example, a molecule with the SMILES string 'Cc1cc(CC2CCC(F)(F)C2)n(Cc2cscn2)c1' with (<x>, <y>, <z>) coordinates for each atom can be output as:\\
    <sdf>\\
    1 30.987 6.655 24.653 C 0\\
    2 31.351 7.912 24.029 C 0\\
    3 30.624 8.775 23.180 C 0\\
    4 31.510 9.726 22.702 C 0\\
    5 31.368 10.813 21.681 C 0\\
    6 30.007 11.245 21.155 C 0\\
    7 28.976 11.750 22.162 C 0\\
    8 27.672 11.818 21.343 C 0\\
    9 27.928 11.007 20.084 C 0\\
    10 26.905 10.209 19.755 F 0\\
    11 28.130 11.857 19.044 F 0\\
    12 29.209 10.235 20.333 C 0\\
    13 32.716 9.446 23.320 N 0\\
    14 34.031 10.014 23.112 C 0\\
    15 35.271 9.181 23.283 C 0\\
    16 35.399 7.840 22.977 C 0\\
    17 36.876 7.257 23.558 S 0\\
    18 37.373 8.844 23.863 C 0\\
    19 36.438 9.775 23.782 N 0\\
    20 32.587 8.443 24.204 C 0\\
    1 2 1 0\\
    2 3 1 0\\
    3 4 2 0\\
    4 5 1 0\\
    5 6 1 0\\
    6 7 1 0\\
    7 8 1 0\\
    8 9 1 0\\
    9 10 1 0\\
    9 11 1 0\\
    9 12 1 0\\
    4 13 1 0\\
    13 14 1 0\\
    14 15 1 0\\
    15 16 2 0\\
    16 17 1 0\\
    17 18 1 0\\
    18 19 2 0\\
    13 20 1 0\\
    20 2 2 0\\
    12 6 1 0\\
    19 15 1 0\\
    </sdf>}
\end{fieldbox}
\end{windowbox}

\begin{windowbox}[width=\linewidth]{Examples of conditions}
\begin{fieldbox}[colback=teal!3]{Pocket description example}
    \textcolor{teal}{
    Format: <pdb\_atom\_name> at (<x>, <y>, <z>)\\
    Chain A:\\
    Residue 109, VAL:\\
    N at (68.271, 2.050, 1.527)\\
    CA at (67.422, 1.308, 2.444)\\
    C at (67.471, -0.161, 2.090)\\
    O at (67.515, -0.532, 0.919)\\
    CB at (65.933, 1.702, 2.355)\\
    CG1 at (65.410, 2.021, 3.730)\\
    CG2 at (65.733, 2.843, 1.411)\\
    ...\\
    Residue 228, PRO:\\
    N at (58.328, 5.229, 16.806)\\
    CA at (57.922, 4.667, 15.514)\\
    C at (56.907, 5.548, 14.808)\\
    O at (56.674, 5.326, 13.600)\\
    CB at (57.330, 3.320, 15.891)\\
    CG at (56.746, 3.565, 17.244)\\
    CD at (57.729, 4.482, 17.928)\\
    OXT at (56.356, 6.445, 15.480)\\
    Chain B:\\
    Residue 57, ALA:\\
    N at (53.770, -1.069, -1.125)\\
    CA at (55.007, -0.739, -0.422)\\
    C at (55.902, 0.060, -1.360)\\
    O at (56.089, -0.311, -2.525)\\
    CB at (55.704, -2.004, 0.009)\\
    ...\\
    Residue 186, TRP:\\
    N at (49.481, 14.438, 11.606)\\
    CA at (49.181, 13.268, 10.788)\\
    C at (48.290, 13.692, 9.624)\\
    O at (47.515, 14.636, 9.741)\\
    CB at (48.472, 12.187, 11.610)\\
    CG at (48.364, 10.864, 10.875)\\
    CD1 at (49.224, 9.798, 10.960)\\
    CD2 at (47.354, 10.487, 9.929)\\
    NE1 at (48.811, 8.787, 10.125)\\
    CE2 at (47.667, 9.181, 9.479)\\
    CE3 at (46.216, 11.126, 9.416)\\
    CZ2 at (46.883, 8.503, 8.540)\\
    CZ3 at (45.437, 10.451, 8.482)\\
    CH2 at (45.777, 9.150, 8.055)}
\end{fieldbox}

\begin{fieldbox}[colback=orange!3]{Anchor fragments description example}
    \textcolor{orange}{
    Format: Element Name (<atomic\_symbol>) within 0.5 Angstroms of (<x>, <y>, <z>)\\
    Carbon (C) within 0.5 Angstroms of (57.658, 3.414, 8.907)\\
    Carbon (C) within 0.5 Angstroms of (58.645, 4.259, 8.129)\\
    Oxygen (O) within 0.5 Angstroms of (59.242, 3.567, 7.049)\\
    Carbon (C) within 0.5 Angstroms of (59.795, 4.701, 9.013)\\
    Oxygen (O) within 0.5 Angstroms of (59.550, 4.399, 10.428)\\
    Carbon (C) within 0.5 Angstroms of (60.108, 6.147, 8.872)\\
    Oxygen (O) within 0.5 Angstroms of (61.217, 6.406, 9.748)\\
    Carbon (C) within 0.5 Angstroms of (58.999, 7.101, 9.267)
    }
\end{fieldbox}

\begin{fieldbox}[colback=olive!3]{Pharmacophore points description example}
    \textcolor{olive}{
    Format: <pharmacophore\_type> within 0.5 Angstroms of (<x>, <y>, <z>)\\
    Hydrogen bond acceptor within 0.5 Angstroms of (54.373, 6.889, 9.107)
    }
\end{fieldbox}

\begin{fieldbox}[colback=violet!3]{Mandatory interactions description example}
    \textcolor{violet}{
    Van der Waals interaction with any atom of residue 74, ILE from chain B
    }
\end{fieldbox}

\end{windowbox}

\newpage
\section{Generated Structures and Unidock Score Distributions}
\label{app:unidock_scores}

\begin{figure}[t]
\centering
\includegraphics[width=0.8\linewidth]{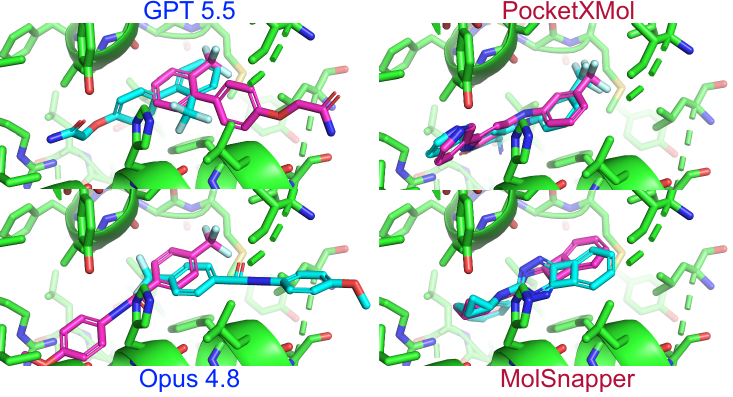}
\caption{Generated structures for \texttt{5TBO} pocket: raw poses in \textcolor{magenta}{magenta}, UniDock-optimized in \textcolor{cyan}{cyan}.}
\label{fig:docked_mols}
\vspace{-10pt}
\end{figure}

Figure \ref{fig:docked_mols} shows several generated examples from different models for the same target. The raw poses from LLMs are significantly different than the final optimized poses, whereas for diffusion models the poses are similar.

\begin{figure}[t]
    \centering
    \includegraphics[width=\textwidth]{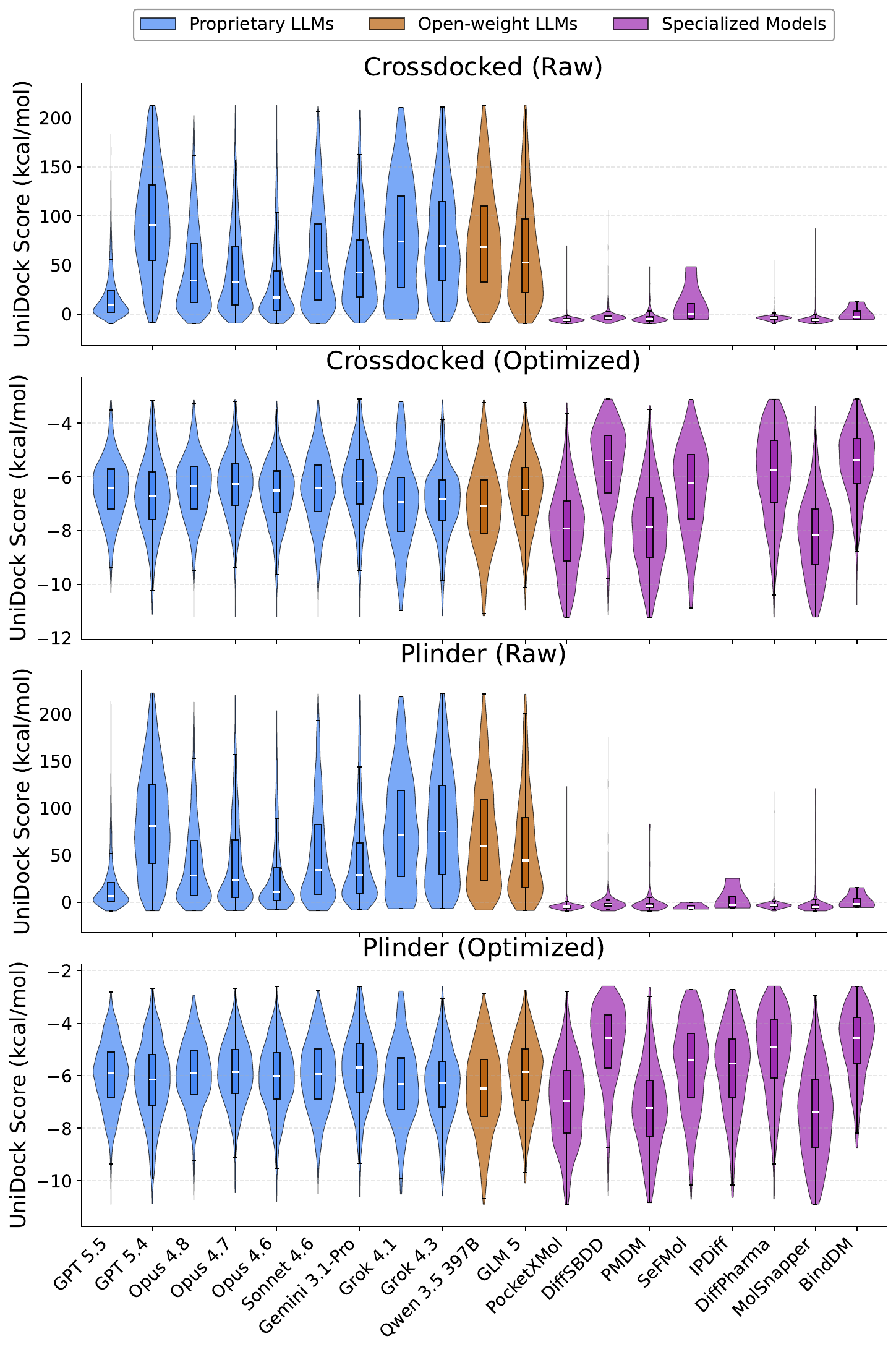}
    \caption{Distributions of Unidock scores for all targets and condition sets, compared between models. 1\% and 99\% outliers are removed.}
    \label{fig:unidock_score_dists}
    \vspace{-10pt}
\end{figure}

Figure \ref{fig:unidock_score_dists} shows the distributions of all Unidock scores from each model for both raw and optimized poses. This figure shows how significantly the LLM-generated poses are optimized compared to diffusion poses, as well as showing the difference in final optimized score between LLMs and Diffusion Models.

\section{Alternative Format: SMILES+XYZ}
\label{app:smiles_xyz}

We additionally benchmarked another 3D ligand output representation, referred to as Enumerated SMILES+XYZ (see Fig.~\ref{fig:smiles_xyz})

\paragraph{Enumerated SMILES+XYZ} 

For this format, we extend the compact SMILES+XYZ textual representation of 3D molecular structures introduced in BindGPT~\cite{zholus2024bindgpt} and nach0-pc~\cite{kuznetsov2024nachopc}. In these works, the molecular graph is first specified using a SMILES string, after which the 3D coordinates of the heavy atoms are provided in the same order as the corresponding atoms appear in the SMILES representation.

We extend this format in a similar to Simplified SDF way by applying an index to each atom in SMILES format, and to its corresponding 3D coordinates. We relax the format fragility by supporting an arbitrary ordering of atomic coordinates as long as coordinate entries cover all heavy atoms in the original SMILES string.

\begin{figure}[b]

\textbf{Enumerated SMILES+XYZ}

{\fontsize{9pt}{10pt}\selectfont
\begin{verbatim}
[c:1]1[c:2][c:3][n:4][c:5][c:6]1|
4 N 1.092 -0.768 0.027|2 C 0.120 1.403 0.009|3 C 1.213 0.555 0.036|1 C -1.123 0.785 -0.028|
6 C -1.216 -0.591 -0.036|5 C -0.087 -1.384 -0.008
\end{verbatim}
}

\caption{Example of Enumerated SMILES+XYZ representation}
\label{fig:smiles_xyz}
\end{figure}

\paragraph{Comparison with Simplified SDF} We benchmarked LLMs on the full-conditioning task (pocket, anchor fragments, pharmacophore points, and mandatory interactions) and found that Simplified SDF consistently outperforms Enumerated SMILES+XYZ (see Table~\ref{tab:all_conditions_format_compare}).

\begin{table*}[th!]
\centering
\setlength{\tabcolsep}{3pt}
\begin{tabular}{lllrrrrrrrrr}
\hline
DS & Model & FMT &
\shortstack{Valid\\Mol} &
\shortstack{PB\\Intra} &
\shortstack{Internal\\Energy} &
\shortstack{PB\\Inter} & 
Pocket &
\shortstack{Uni-\\Dock} & PH4. & 
\shortstack{3D\\Anch.} & 
\shortstack{Mand.\\Int.} \\
\hline
\multirow{12}{*}{\rotatebox[origin=c]{90}{CrossDocked2020}}
& \multirow{2}{*}{GPT-5.4} & SDF    & 0.979 & 0.420 & 0.530 & 0.308 & 0.608 & -5.320 & 0.730 & 0.977 & 0.242 \\
& & SMI & 0.808 & 0.091 & 0.230 & 0.342 & 0.665 & -5.183 & 0.742 & 0.660 & 0.271 \\
\cline{2-12}
& \multirow{2}{*}{Sonnet 4.6} & SDF    & 0.678 & 0.405 & 0.530 & 0.439 & 0.441 & -4.844 & 0.519 & 0.678 & 0.175 \\
&  & SMI & 0.467 & 0.282 & 0.344 & 0.362 & 0.290 & -4.594 & 0.430 & 0.457 & 0.124 \\
\cline{2-12}
& \multirow{2}{*}{Opus 4.6} & SDF    & 0.588 & 0.400 & 0.504 & 0.230 & 0.466 & -5.465 & 0.436 & 0.583 & 0.228 \\
& & SMI & 0.440 & 0.251 & 0.341 & 0.210 & 0.363 & -5.368 & 0.432 & 0.431 & 0.175 \\
\cline{2-12}
& \multirow{2}{*}{Gemini 3.1 Pro} & SDF    & 0.965 & 0.720 & 0.885 & 0.543 & 0.827 & -5.246 & 0.733 & 0.965 & 0.487 \\
&  & SMI & 0.969 & 0.718 & 0.891 & 0.542 & 0.804 & -5.129 & 0.893 & 0.969 & 0.480 \\
\cline{2-12}
& \multirow{2}{*}{Qwen 3.5} & SDF    & 0.898 & 0.339 & 0.496 & 0.640 & 0.574 & -4.639 & 0.670 & 0.891 & 0.255 \\
& & SMI & 0.823 & 0.282 & 0.398 & 0.667 & 0.442 & -4.278 & 0.709 & 0.787 & 0.191 \\
\cline{2-12}
& \multirow{2}{*}{GLM 5} & SDF    & 0.400 & 0.270 & 0.274 & 0.344 & 0.226 & -4.275 & 0.350 & 0.399 & 0.119 \\
& & SMI & -- & -- & -- & -- & -- & -- & -- & -- & -- \\

\hline

\multirow{12}{*}{\rotatebox[origin=c]{90}{PLINDER}}
& \multirow{2}{*}{GPT-5.4} & SDF    & 0.990 & 0.386 & 0.448 & 0.327 & 0.479 & -4.732 & 0.790 & 0.988 & 0.317 \\
&  & SMI & 0.812 & 0.057 & 0.133 & 0.354 & 0.507 & -4.609 & 0.768 & 0.614 & 0.240 \\
\cline{2-12}
& \multirow{2}{*}{Sonnet 4.6} & SDF    & 0.685 & 0.440 & 0.554 & 0.396 & 0.317 & -4.246 & 0.578 & 0.683 & 0.230 \\
&  & SMI & 0.434 & 0.234 & 0.299 & 0.279 & 0.176 & -3.913 & 0.404 & 0.416 & 0.158 \\
\cline{2-12}
& \multirow{2}{*}{Opus 4.6} & SDF    & -- & -- & -- & -- & -- & -- & -- & -- & -- \\
& & SMI & 0.489 & 0.315 & 0.388 & 0.220 & 0.277 & -4.463 & 0.473 & 0.483 & 0.218 \\
\cline{2-12}
& \multirow{2}{*}{Gemini 3.1 Pro} & SDF    & 0.978 & 0.679 & 0.895 & 0.517 & 0.707 & -4.771 & 0.786 & 0.978 & 0.594 \\
& & SMI & 0.960 & 0.659 & 0.877 & 0.495 & 0.640 & -4.629 & 0.909 & 0.960 & 0.584 \\
\cline{2-12}
& \multirow{2}{*}{Qwen 3.5} & SDF    & 0.869 & 0.319 & 0.450 & 0.598 & 0.400 & -4.138 & 0.685 & 0.863 & 0.309 \\
& & SMI & 0.770 & 0.208 & 0.299 & 0.584 & 0.250 & -3.741 & 0.677 & 0.713 & 0.204 \\
\cline{2-12}
& \multirow{2}{*}{GLM 5} & SDF    & 0.420 & 0.267 & 0.281 & 0.289 & 0.152 & -3.753 & 0.382 & 0.414 & 0.172 \\
& & SMI & -- & -- & -- & -- & -- & -- & -- & -- & -- \\

\hline
\end{tabular}
\caption{3D Quality Metrics and Condition Successes for Simplified SDF and Enumerated SMILES+XYZ formats.}
\label{tab:all_conditions_format_compare}
\end{table*}

\end{document}